\documentclass{article}
\usepackage{amsmath,epsfig}
\usepackage[preprint]{spconfa4}

\usepackage[small]{caption}
\usepackage{graphicx}
\usepackage{cite}
\usepackage{caption}
\usepackage{subfigure}
\usepackage{amsmath}

\let\OLDthebibliography\thebibliography
\renewcommand\thebibliography[1]{
  \OLDthebibliography{#1}
  \setlength{\parskip}{0pt}
  \setlength{\itemsep}{0pt plus 0.3ex}
}

\begin{document}\sloppy

\def\x{{\mathbf x}}
\def\L{{\cal L}}

\title{Fine-Grained Texture Identification for Reliable Product Traceability
\vspace{-8pt}}
%
\name{Junsong Wang, Yubo Li, Zhiyong Chang, Haitao Yue, Yonghua Lin \vspace{-12pt}}
\address{V-Origin Technology, Beijng, China \\ \{junosng.wang, yubo.li, zhiyong.chang, haitao.yue, yonghua.lin\}@easy-visible.com \vspace{-16pt}}

\maketitle{}

\begin{abstract}
Texture exists in lots of the products, such as wood, beef and compression tea. These abundant and stochastic texture patterns are significantly different between any two products. Unlike the traditional digital ID tracking, in this paper, we propose a novel approach for product traceability, which directly uses the natural texture of the product itself as the unique identifier. A texture identification based traceability system for Pu'er compression tea is developed to demonstrate the feasibility of the proposed solution. With tea-brick images collected from manufactures and individual users, a large-scale dataset has been formed to evaluate the performance of tea-brick texture verification and searching algorithm. The texture similarity approach with local feature extraction and matching achieves the verification accuracy of 99.6\% and the top-1 searching accuracy of 98.9\%, respectively. 



\end{abstract}


\section{Introduction}

Product traceability is the capability of tracking every aspect of manufacturing and distributing process. Lots of technologies have been proposed to solve this challenging problem. In some traceability systems, each product is bound with either a unique quick response (QR) code or near field communication (NFC) chip, so the whole manufacturing, warehousing, distribution, in-store selling and logistics could be well traced. However, all these digital ID based solutions can not completely solve the counterfeiting problem, since both the QR code and NFC chip are easy to be duplicated, or recycled and attached in another faked product.

\begin{figure}[htbp]
  \centering
  \includegraphics[width=0.4\textwidth]{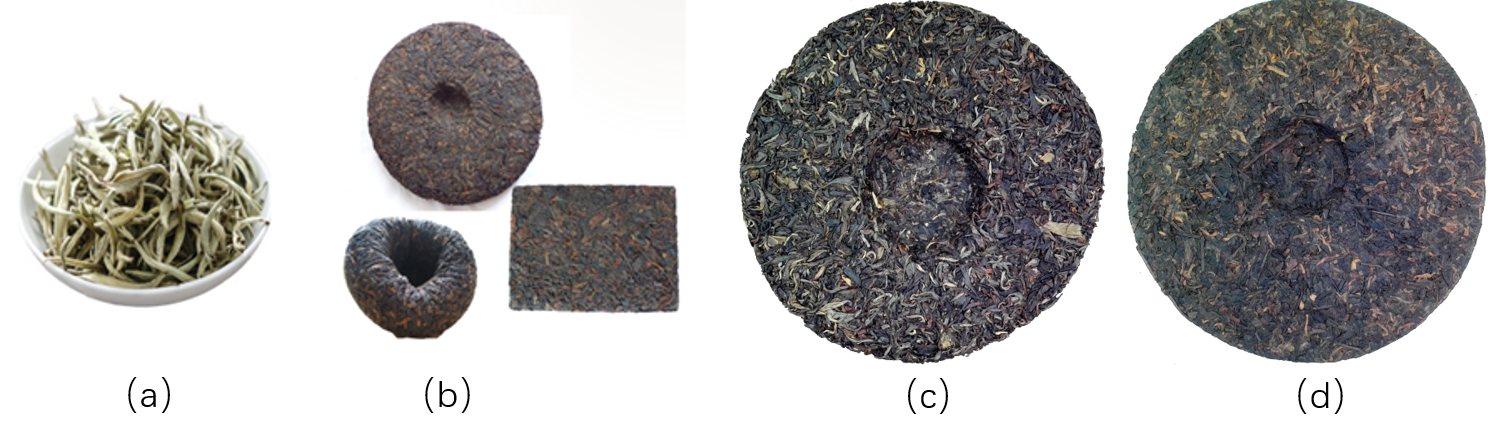}
  \vspace{-8pt}
  \caption{Some examples of Pu'er tea. (a) loose tea leaves, (b) several types of tea-brick. (c) and (d) show the detailed texture property of raw tea and ripe tea (raw tea with fermentation), respectively.}
  \label{fig:tea-brick}
  \vspace{-12pt}
\end{figure}

The best approach for reliable traceability is to extract some natural and unique information from the product itself, which is impossible to be duplicated or counterfeited. Actually, natural texture exists in lots of the products. Some of them are born with different natural texture patterns, such as wood, jade and meat. Some of them are generated during production, such as compression tea and cork of wine. Although texture analysis has been already discussed in some literatures for remote sensing, material classification, etc., the fine-grained texture identification problem, which in concept is very similar to facial recognition, was never addressed. In this paper, we develop a system TI-Trace to do texture identification by machine automatically, and successfully apply it to product traceability domain. We select the Pu'er tea as the first use case because 1) it is facing a serious counterfeiting problem in the market, 2) the texture of the tea-brick formed during the compression process is sufficient enough for identification as illustrated in Fig.\ref{fig:tea-brick}.

\section{Related Work}\label{related-work}
Texture analysis has been widely used in a broad range of applications, such as texture classification based remote sensing and segmentation based biomedical imaging. A series of feature extraction methods were discussed, including statistical approaches, transform-based approaches, etc. They mainly focus on the classification from different type of materials. However, the fine-grained texture identification among single type of material/product is never discussed. 

Actually, the texture identification task can also be considered as a image retrieval problem, following a standard framework where an image is embedded with a bag-of-visual-words (BoW) and ranked using standard term-frequency inverse-document-frequency (TF-IDF) weighting scheme \cite{bow}. This embedding technology significantly reduces the image similarity computation and is applicable to large-scale retrieval. The evaluation results over a number of standard benchmark datasets (Oxford Buildings 105k and Paris 6k) show the retrieval accuracy is too low and to be used for product identification.

Image matching can identify the similar content between two images with a classical pipeline of local feature extraction, feature matching and geometric verification \cite{iamge-matching}. The literature in \cite{iamge-matching-cnn} introduces convolutional neural network (CNN) to extract feature descriptors instead of SIFT and achieves performance improvement in descriptor matching. Although in the large scale identification task, the feature descriptors could be calculated offline, the matching process and geometric verification for each pair of images, which usually involve nearest neighbor (NN, $O(n^2)$ complexity) and random sample consensus (RANSAC) algorithm \cite{RANSAC} respectively, are still time intensive.

The most promising approach to solve the fine-grained image identification problem is the deep metric learning, which uses a deep CNN for feature extraction and a dedicated loss function, such as contrastive loss and triplet loss. The image is eventually embedded into a shot vector, and the similarity can be obtained by calculating their Euclidean or cosine distance. The deep metric learning achieves the state-of-the-art performance in facial recognition and person re-identification task. However, it needs a huge number of labeled data for training. For example, in facial recognition, millions of labeled facial images are used for training. 

\section{TI-Trace System Overview}\label{system-overview}
\begin{figure}[htbp]
  \centering
  \includegraphics[width=0.4\textwidth]{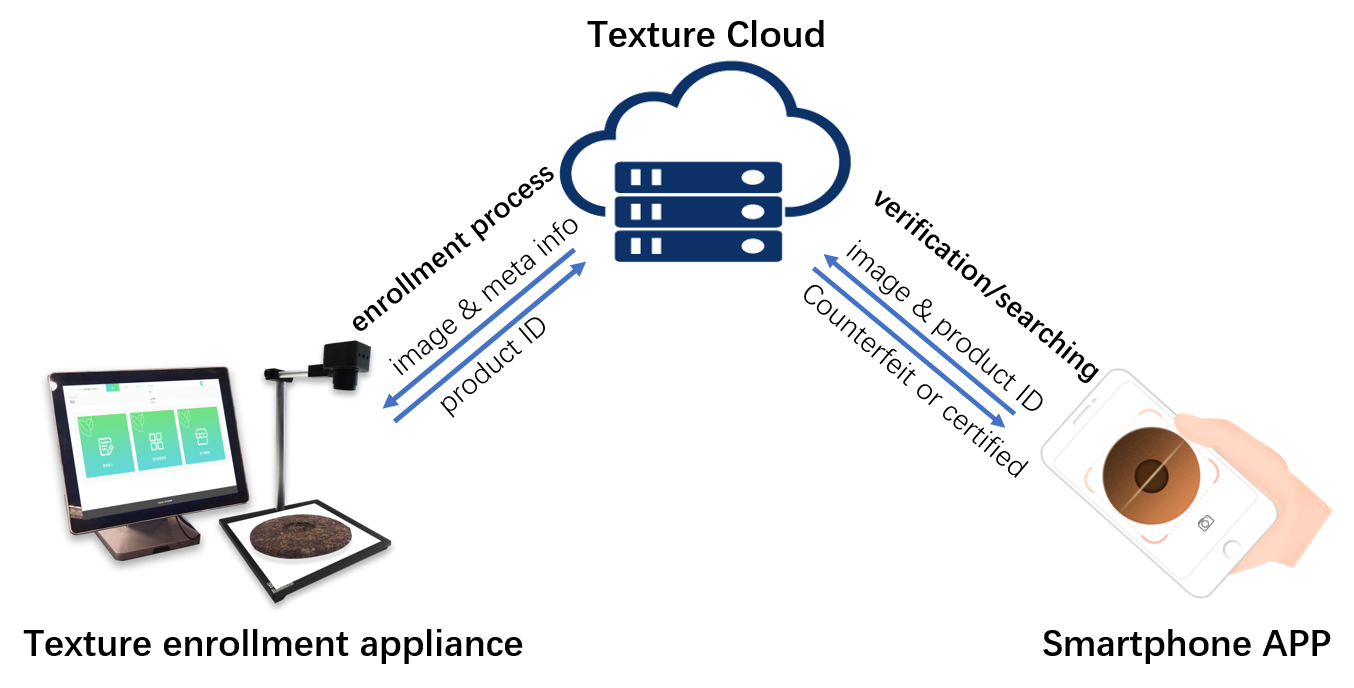}
  \vspace{-12pt}
  \caption{TI-Trace system architecture overview}
  \label{fig:system}
  \vspace{-14pt}
\end{figure}

Similar to the facial identification, our TI-Trace system also has the capability of texture enrollment, verification and large-scale searching. As illustrated in Fig. \ref{fig:system}, the whole system consists of the following three major components.


\textbf{Texture enrollment appliance}: This assembled device is well designed for the manufacture's workers to capture the product's texture image and enroll to our TI-Trace system in a convenient and efficient way. An industrial camera is installed inside a simple camera shed. Two LED tubes are mounted on the top of the camera shed to guarantee the illumination quality since the natural light in the factory is usually insufficient. The industrial computer will capture the textural image through the camera, do image pre-processing, such as localization, cropping and resizing, and finally, enroll the textural image to the backend texture cloud together with the product's metadata, such as manufacture name, batch code, category, etc. A unique ID is allocated by the texture cloud after successful enrollment.

\textbf{Texture cloud}: The cloud based tea-brick texture management system provides 1) some basic interfaces for textural image enrollment and essential product management functions, 2) texture verification by calculating the similarity between the enrolled image and the uploaded one from end-user, 3) searching function to retrieve an enrolled product only by uploading a textural image.

\textbf{Light-weight APP on the smartphone} is developed for customers to perform product verification and searching using its natural texture. In this APP, by scanning the attached QR code on the product packaging, customer can get all the related information, such as manufacture, place of origin, selling and logistics. For further verification with texture, the customer can take a photo of the product, and upload it to the texture cloud to check whether it is authentic. The customer could also retrieve product information as well as check authenticity by uploading a textural image to perform the global searching even if the attached QR code is lost.

\section{Tea-brick Texture Dataset and Identification Task}
We take one of the most popular compression tea, Pu'er tea, as the first use case, and build a dataset for algorithm evaluation. Each tea-brick consists of more than ten thousand of tea leaves and has sufficient texture information as shown in Fig.\ref{fig:tea-brick}(c)(d). 

\begin{figure}[htbp]
  \centering
  \includegraphics[width=0.42\textwidth]{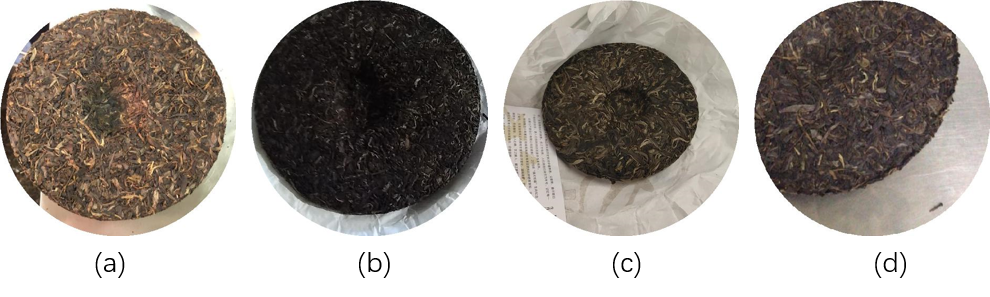}
  \vspace{-12pt}
  \caption{Uploaded images from the customer. (a)ideal sample, (b)insufficient illumination, (c)bad viewpoint and position, (d)serious occlusion.}
  \label{fig:bad_example}
  \vspace{-12pt}
\end{figure}

The tea-brick texture dataset is divided into two subsets. The gallery dataset is captured from the manufacture product line with our texture enrollment appliance in constrained environment. The texture is very clear without any occlusion as shown in Fig.\ref{fig:tea-brick}(c)(d). The query dataset is captured by the customers with different types of smartphones, and the capture conditions is very diverse as shown in Fig.\ref{fig:bad_example}, i.e., capturing viewpoints (the angle to capture the image), occlusions and illuminations.

The gallery dataset contains 30,000 samples in total, which is a subset of the enrolled images captured from more than sixty different manufactures and hundreds of product batches. The query dataset has two parts with a training subset of 2276 images and a test subset of 358 images. Since the query dataset is still somewhat out of quality control by different shot behaviors, the images are first checked by a quality evaluation network using ResNet18, and further manually checked with reasonable illumination, viewpoint and limited occlusion. The better image quality definitely helps to improve the verification and searching accuracy, but user experience may be affected since the customer may need to take several shots to obtain a satisfied photo. Therefore, we also provide a hard test dataset (64 images) with extremely low illumination, large viewpoint and serious occlusion.


We use the gallery dataset and test query dataset to generate the verification dataset. For each image in test query dataset, we generate one positive pair and one negative pair. The positive pair is from gallery-query pair while the negative one is randomly selected from the gallery set. The verification dataset totally has 716 test pairs. The test query dataset is also used to evaluate the searching performance from the gallery set.

\textbf{Remark}: Although this texture identification and facial recognition tasks are very similar, both of which are open-set (test set and training set classes are disjoint) recognition tasks and require extremely high accuracy in real applications, it still has very unique challenges, 1) there is no landmark in the tea-brick's texture, which can not be well aligned before the feature extraction; 2) the query set with diverse illumination and viewpoint is relatively too small to generate as many positive pairs as in large-scale facial dataset, which will be the big barrier for advanced metric learning.


\section{Performance Analysis}
\subsection{Algorithm study}
Firstly, we study some potential approaches derived from other researches, and their accuracies on the verification dataset are shown in Table.\ref{algorithm-se}. For the BoW approach, parts of the gallery set images and all the training query images are used to extract totally 5 millions SIFT descriptors,followed by k-means clustering approach to generate 5,000 centroids, which are considered as the visual words. In deep metric learning approach, we select ResNet50 as the backbone, and supervised with the triplet loss. 

For the image matching approach, we select three popular local feature extractors, scalable invariant feature transform (SIFT) \cite{sift}, speeded-up robust features (SURF) \cite{surf}, and oriented FAST and rotated BRIEF (ORB) \cite{orb}. NN algorithm is adopted to find the good matches using the nearest neighbor ratio test. It firstly calculates the distance between each pair of feature descriptors, and then check the distance ratio between the nearest neighbor and the second-nearest neighbor. The pair is considered as a good match only if the ratio is less than a threshold, which called distance ratio threshold (DRT) in the following. In this experiment, we don't restrict the number of the keypoints in image matching, and set DRT to 0.75. Finally, RANSAC algorithm based geometric verification is performed to remove mismatched keypoint pairs. The two textural images are judged as the same product only if the total matched pair is above a threshold, which is called matched pair threshold (MPT) in the following.


From Table.\ref{algorithm-se}, we can see that the image matching approach with local feature achieves a pretty good accuracy. Comparing to SIFT extractor, the SURF and ORB extractor suffer an obvious performance degradation of 2x and 5x in terms of error rate. Therefore, considering the demand of extremely high accuracy in product identification, we eventually select SIFT as the local feature extractor although the computation cost is high.

\begin{table}[htbp]
\footnotesize
  \centering
  \caption{Comparison of BoW, image matching and metric learning.}
   \vspace{-8pt}
  \label{algorithm-se}
  \begin{tabular}{|c|c|}
  \hline
    \textbf{methods} & \textbf{verification accuracy} \\
    \hline
    BoW + tidif ranking  & 73.0\%  \\
    \hline
    ResNet50+Triplet Loss &   80.1\% \\
    \hline
    Image Matching, SIFT &  99.6\%  \\
    \hline
    Image Matching, SURF &  99.2\%  \\
    \hline
    Image Matching, ORB &  97.9\%  \\
    \hline
  \end{tabular}
  \vspace{-12pt}
\end{table}

\subsection{Performance with different configurations in SIFT}

Table.\ref{tbl:roc_with_ratio} shows the accuracy over DRT from 0.6 to 0.8 with 600 keypoints. We can observe that the accuracy is improved from 97.1\% to 99.0\%, which has nearly 3x improvement in terms of error rate. We can also find DRT has little effect on the final accuracy if it is in the range from 0.7 to 0.8. In other hand, a bigger ratio will increase the probability of geometric verification for negative pairs, which has serious negative impact on searching speed as discussed in section \ref{searching}.

\begin{table}[htbp]
  \footnotesize
  \centering
  \caption{Accuracy with different DRT, 600 keypoints.}
  \vspace{-8pt}
  \label{tbl:roc_with_ratio}
  \begin{tabular}{|c|c|c|c|c|c|}
  \hline
    DRT & 0.60 & 0.65 & 0.70 & 0.75 & 0.80  \\
    \hline
    Accuracy & 96.79\%  & 97.91\%  & 98.74\%  & 98.99\% & 99.02\%  \\
    \hline
  \end{tabular}
   \vspace*{-12pt}
\end{table}

\begin{table}[htbp]
  \footnotesize
  \centering
  \caption{Accuracy with different number of keypoints, DRT is 0.75.}
  \vspace{-8pt}
  \label{tbl:roc_with_kp}
  \begin{tabular}{|c|c|c|c|c|c|}
  \hline
    keypoints & 200 & 400 & 600 & 800 & 100  \\
    \hline
    Accuracy & 93.30\%  & 97.77\%  & 98.88\%  & 99.30\% & 99.58\%  \\
    \hline
  \end{tabular}
   \vspace*{-8pt}
\end{table}

We also investigated the effect of different number of extracted keypoints, which is illustrated in Table.\ref{tbl:roc_with_kp}. The DRT is fixed to 0.75. The best accuracy reaches to 99.6\% in the case of extracting 1000 keypoints. The more keypoints we extracted, the better accuracy we can achieved, but we need afford more computation and storage resources.

\subsection{Performance on the hard verification dataset}
Firstly, we consider the effect of serious occlusion using an augmented dataset by cropping the image from the test query set. Half of the cropped patches only keep a quarter of their original one. The result in Table.\ref{match:hard} is very encouraging. Even with serious occlusion, the accuracy only drops 1.8\% in the best case of adopting 1000 keypoints, which indicates the algorithm is not very sensitive to occlusion. Even the tea-brick is broken, or a small fraction remains, the remainder can still be used for verification.

\begin{table}[htbp]
\footnotesize
  \centering
  \caption{Accuracy on serious occlusion. The DRT is 0.75.}
   \vspace{-8pt}
  \label{match:hard}
  \begin{tabular}{|c|c|c|c|c|c|}
  \hline
    \# of keypoints & 800 & 1000  \\
    \hline
    Accuracy (normal)  & 99.3\%  & 99.6\%  \\
    \hline
    Accuracy (w/ occlusion) & 97.3\%(2.0\% $\downarrow$)  & 97.8\%(1.8\% $\downarrow$)  \\
    \hline
  \end{tabular}
  \vspace*{-12pt}
\end{table}

Secondly, we refer to the hard test dataset. Most of the images suffer from very low illumination and the viewpoint is up to $60^{\circ}$. The true positive rate (TPR) drops significantly to 21.8\%, which indicates the algorithm is very sensitive to illumination and viewpoint. Luckily, as we will discuss in section \ref{searching}, the false positive rate (FPR) drops exponentially with the increase of MPT, we could slightly increase MPT and involve multiple pre-processing and matching technologies to boost the TPR without increasing FPR. Usually, the TPR is much more important since it will introduce unexpected dispute if the system mistakenly judges the certified product as a counterfeit one. The boosting strategies includes, 1) maximal keypoints of 1600, 2) multiple resolutions with $448 \times 448$ and $672 \times 672$, 3) brightness equalization in prep-processing, 4) cross checking (not mutual validation) in nearest neighbor matching. By setting MPT=9, the TPR is significantly boosted to 61.9\% and the FPR is still kept. 

Actually, the SIFT features and correspondences matching is not stable, and some slight changes in the image will make the matching result significantly different. Therefore, for positive pair, we need to keep lots of keypoints and multiple pre-processing strategies to boost the probability of matching, while for negative pair, this phenomenon doesn't exist because of the geometric verification using homography.     

\subsection{Performance of Texture Searching}\label{searching}
In this section, we will evaluate the searching performance. By balancing the accuracy and computation complexity of NN matching, we use the typical setting of 768 keypoints, which makes the descriptor matrix GPU memory friendly.

\begin{figure}[htbp]
  \centering
  \includegraphics[width=0.45\textwidth]{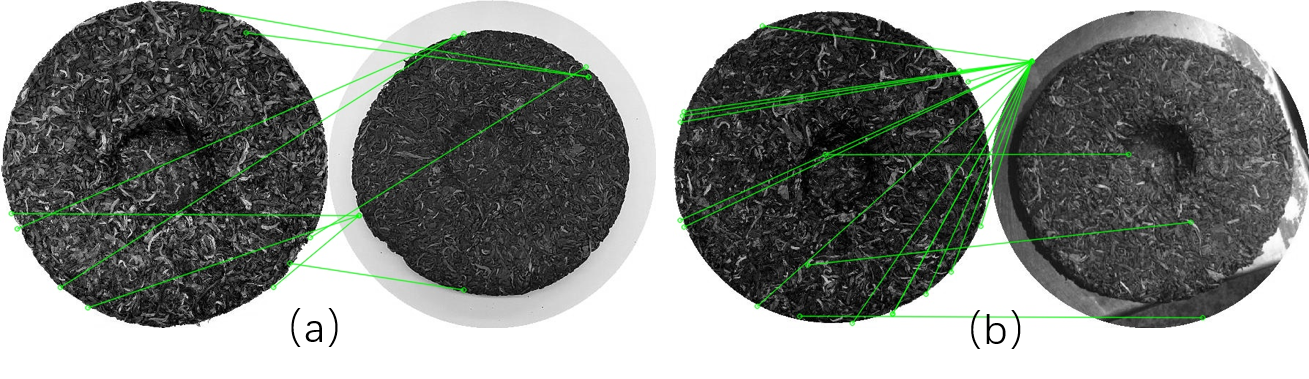}
  \vspace{-12pt}
  \caption{Examples of error matching, (a) edge matching, (b) one to many matching.}
  \label{fig:edge_match}
  \vspace{-12pt}
\end{figure}

We firstly achieved the best top-1 accuracy of 97.48\% as illustrated in Table.\ref{search-top1}. By carefully investigating the error example and visualizing the wrongly matched correspondences, we found that the edge features (keypoint near to the edge) of the tea-brick are easier to be mismatched as shown in Fig.\ref{fig:edge_match}. Most of the error correspondences are similar to Fig.\ref{fig:edge_match}(a), and can be easily handled by homography verification. Some of them are caused by one to many matching, which can also be suppressed by only keeping one correspondence. 

However, homography verification is a big barrier for searching acceleration. From the large-scale searching point of view, since the descriptors could be extracted off-line, the major computation is the NN matching and homography verification. The NN matching could be easily accelerated by GPU because of its natural parallel pattern, while the homography verification is not GPU friendly. If a large part of the matching process involves the homography verification, the searching speed will be significantly affected.  


\begin{table}[htbp]
  \footnotesize
  \centering
  \caption{Top-1 searching accuracy improvement with edge feature exclusion. The DRT is set to 0.7/0.75. The number of keypoints is 768}
  \label{search-top1}
  \vspace{-8pt}
  \begin{tabular}{|c|c|c|}
  \hline
    DRT & 0.70 & 0.75  \\
    \hline
    Acc. w/o edge exclusion & 97.20\%  & 97.48\%   \\
    \hline
    Acc. w/ edge exclusion & 98.32\% (1.12\% $\uparrow$ ) & 98.85 \% (1.43\% $\uparrow$ ) \\
    \hline
  \end{tabular}
  \vspace*{-12pt}
\end{table}

To overcome this issue, we use an alternative way by performing edge detection for both gallery and query images during the feature extraction. All the edge features (within the distance of 5 pixels to the detected edge) will be removed. Due to the diversity of the image in the query set, the edge detection is not always perfect. If we directly remove the edge features before the NN matching, the remained unrecognized edge features still can not be well suppressed because we can not benefit from the intrinsic auto-suppression property of NN matching. Therefore, we should do the edge feature exclusion after the NN matching. With this simple approach, we can eliminate most of the error edge correspondences. As shown in Fig.\ref{fig:edge_match}, nearly all the mismatched correspondences could be removed without using homography. Thus, the mismatched correspondences in most of the negative pairs during the searching is below the threshold and no further homography verification is required, The homography computation is saved with a larger margin, and won't obviously increase the searching time.

As shown in Table.\ref{search-top1}, the final accuracy is improved by 1.43\% for DRT $=0.75$, and the average searching speed with Nvidia V100 GPU acceleration is increased by 6.2x since very few negative pairs need further homography verification.

\section{Conclusion} 
In order to completely solve the counterfeiting problem for reliable product traceability, we proposed to use its natural texture information as the unique feature for identification. This approach creates a strong and natural correlation between the digital ID and the physical product, which is indestructible and unforgeable. We have achieved pretty good accuracy in tea-brick's texture identification and demonstrated the feasibility of our traceability system. 

\small
\bibliographystyle{IEEEbib}

\bibliography{ref}

\begin{thebibliography}{1}

\bibitem{bow}
J.~Sivic and A.~Zisserman,
\newblock ``Video google: A text retrieval approach to object matching in
  videos,''
\newblock {\em ICCV}, 2003.

\bibitem{iamge-matching}
J.~Ma, X.~Jiang, A.~Fan, J.~Jiang, and J.~Yan,
\newblock ``Image matching from handcrafted to deep features: A survey,''
\newblock {\em IJCV}, 2020.

\bibitem{iamge-matching-cnn}
P.~Fischer, T.~Dosovitskiy, and T.~Brox,
\newblock ``Descriptor matching with convolutional neural networks: a
  comparison to sift,''
\newblock {\em arXiv preprint arXiv:1405.5769}, 2014.

\bibitem{RANSAC}
M.~A. Fischler and R.~C. Bolles,
\newblock ``Random sample consensus: A paradigm for model fitting with
  applications to image analysis and automated cartography,''
\newblock in {\em Communications of the ACM}, 1981.

\bibitem{sift}
D.~G. Lowe,
\newblock ``Distinctive image features from scale-invariant keypoints,''
\newblock in {\em International Journal of Computer Vision}, 2004.

\bibitem{surf}
H.~Bay, T.~Tuytelaars, and L.~V. Gool,
\newblock ``Speeded-up robust features (surf),''
\newblock in {\em Computer vision and image understanding}, 2008.

\bibitem{orb}
E.~Rublee, V.~Rabaud, K.~Konolige, and G.~Bradski,
\newblock ``Orb:and efficient alternative to sift or surf,''
\newblock in {\em International Conference on Computer Vision (ICCV)}, 2011.

\end{thebibliography}

\end{document}